\documentclass{article}

\usepackage[preprint]{neurips_2024}

\usepackage[utf8]{inputenc}
\usepackage[T1]{fontenc}
\usepackage{hyperref}
\usepackage{url}
\usepackage{booktabs}
\usepackage{amsfonts}
\usepackage{amsmath}
\usepackage{nicefrac}
\usepackage{microtype}
\usepackage{xcolor}
\usepackage{graphicx}
\usepackage{natbib}
\usepackage{float}  
\usepackage{multirow}  

\title{Detecting and Steering LLMs' Empathy in Action}

\author{%
  Juan P. Cadile \\
  Department of Philosophy \\
  University of Rochester \\
  \texttt{jcadile@ur.rochester.edu}
}

\begin{document}

\maketitle

\begin{abstract}
We investigate empathy-in-action---the willingness to sacrifice task efficiency to address human needs---as a linear direction in LLM activation space. Using contrastive prompts grounded in the Empathy-in-Action (EIA) benchmark, we test detection and steering across Phi-3-mini-4k (3.8B), Qwen2.5-7B (safety-trained), and Dolphin-Llama-3.1-8B (uncensored).

\textbf{Detection:} All models show AUROC 0.996--1.00 at optimal layers. Uncensored Dolphin matches safety-trained models, demonstrating empathy encoding emerges independent of safety training. Phi-3 probes correlate strongly with EIA behavioral scores ($r=0.71$, $p<0.01$). Cross-model probe agreement is limited (Qwen: $r=-0.06$, Dolphin: $r=0.18$), revealing architecture-specific implementations despite convergent detection.

\textbf{Steering:} Qwen achieves 65.3\% success with bidirectional control and coherence at extreme interventions ($\alpha=\pm20$). Phi-3 shows 61.7\% success with similar coherence. Dolphin exhibits asymmetric steerability: 94.4\% success for pro-empathy steering but catastrophic breakdown for anti-empathy (empty outputs, code artifacts).

\textbf{Implications:} The detection-steering gap varies by model. Qwen and Phi-3 maintain bidirectional coherence; Dolphin shows robustness only for empathy enhancement. Safety training may affect steering \emph{robustness} rather than preventing manipulation, though validation across more models is needed.
\end{abstract}

\section{Introduction}

Behavioral empathy benchmarks such as Empathy-in-Action (EIA)~\citep{eia2024} provide rigorous tests of empathic reasoning but are expensive to run. Activation probes offer a promising alternative: cheap, online monitoring directly from model internals~\citep{marks2023geometry,zou2023representation}.

\textbf{We define ``empathy-in-action'' as the willingness to divert from a requested goal to address an observed need, even when doing so hinders or derails the primary objective.} This definition captures the core tension: empathy requires recognizing distress and choosing compassionate action despite task-efficiency tradeoffs.

We test this across three models: Phi-3-mini-4k-instruct (3.8B), Qwen2.5-7B-Instruct (safety-trained), and Dolphin-Llama-3.1-8B (uncensored). This diversity allows us to examine how safety training affects empathy encoding and manipulation.

However, a critical question remains: \textbf{do probes capture causal mechanisms or merely correlational features?} A probe that successfully \emph{detects} empathic text may not enable \emph{steering} empathic behavior if it captures surface correlates rather than underlying reasoning.

We investigate this detection-vs-steering gap through four research questions:
\begin{enumerate}
    \item Can empathy be detected as a linear direction in activation space?
    \item Do empathy probes generalize across model architectures?
    \item Do probe projections correlate with behavioral outcomes?
    \item Can we steer empathic behavior by adding the probe direction?
\end{enumerate}

\textbf{Key findings:} Detection succeeds (AUROC 0.96--1.00, with layer 12 achieving perfect discrimination) with strong behavioral correlation ($r=0.71$). Steering reveals model-specific patterns: Qwen2.5-7B (7B, safety-trained) achieves 65.3\% success with bidirectional control; Dolphin-Llama-3.1-8B (8B, uncensored) shows 94.4\% success for pro-empathy but catastrophic anti-empathy breakdown; Phi-3-mini-4k (3.8B) achieves 61.7\% success with coherence similar to Qwen. Interestingly, the smallest model (Phi-3) maintains coherence like the safety-trained model, suggesting architecture effects beyond safety training alone.

\section{Related Work}

\paragraph{Linear representations and probes.}
The linear representation hypothesis~\citep{elhage2022toy,park2023linear} posits that high-level concepts encode as linear directions in activation space. Recent work validates this: \citet{zou2023representation} extracted ``honesty'' directions, \citet{marks2023geometry} analyzed refusal mechanisms, and \citet{turner2023activation} demonstrated steering through activation addition. Our work extends this to \emph{empathy}, a complex socio-emotional concept.

\paragraph{Behavioral empathy benchmarks.}
\citet{eia2024} introduced Empathy-in-Action, testing whether agents sacrifice task objectives to help distressed users. EIA scenarios create \textbf{task-objective conflicts} (efficiency vs compassion), enabling rigorous behavioral tests but potentially confounding probe extraction.

\paragraph{Steering limitations.}
While activation steering shows promise~\citep{turner2023activation,li2024inference}, limitations exist: \citet{jain2024mechanistically} found safety training resists steering, and \citet{huang2023catastrophic} showed inconsistent effects in complex scenarios. We contribute evidence that \emph{task-objective conflicts specifically} impede additive steering.

\section{Method}

\subsection{Contrastive Dataset Generation}

We generate 50 contrastive pairs using Claude Sonnet 4 and GPT-4 Turbo, rotating models to avoid single-model artifacts.

\paragraph{Scenarios.} Five EIA scenarios (Food Delivery, The Listener, The Maze, The Protector, The Duel), each presenting task-empathy conflicts (e.g., ``maximize points'' vs ``help distressed user'').

\paragraph{Prompts.} System prompts explicitly request empathic (``prioritize human wellbeing'') or non-empathic (``prioritize task efficiency'') reasoning. See Appendix~\ref{app:prompts} for full prompts.

\paragraph{Split.} 35 training pairs, 15 test pairs (70/30 split).

\subsection{Probe Extraction}

We extract probes from all three models---Phi-3-mini-4k-instruct~\citep{abdin2024phi3} (3.8B), Qwen2.5-7B-Instruct, and Dolphin-Llama-3.1-8B---using mean difference:
\begin{equation}
\mathbf{d}_{\text{emp}} = \frac{\mathbb{E}[\mathbf{h}_{\text{emp}}] - \mathbb{E}[\mathbf{h}_{\text{non}}]}{\|\mathbb{E}[\mathbf{h}_{\text{emp}}] - \mathbb{E}[\mathbf{h}_{\text{non}}]\|}
\end{equation}
where $\mathbf{h} \in \mathbb{R}^{d}$ are mean-pooled activations from layers $\ell \in \{8, 12, 16, 20, 24\}$.

\paragraph{Validation.} AUROC, accuracy, and class separation on 15 held-out pairs.

\subsection{Behavioral Correlation}

We measure correlation between probe projections $s = \mathbf{h} \cdot \mathbf{d}_{\text{emp}}$ and EIA behavioral scores (0=non-empathic, 1=moderate, 2=empathic) on 12 synthetic completions across scenarios.

\subsection{Activation Steering}

During generation, we add scaled probe direction:
\begin{equation}
\mathbf{h}' = \mathbf{h} + \alpha \cdot \mathbf{d}_{\text{emp}}
\end{equation}
with $\alpha \in \{-20, -10, -5, -3, -1, 0, 1, 3, 5, 10, 20\}$ for Phi-3, $\alpha \in \{-20, -10, -5, -3, 0, 3, 5, 10, 20\}$ for Qwen, and $\alpha \in \{-10, -5, -3, 0, 3, 5, 10\}$ for Dolphin (limited due to coherence breakdown), temperature 0.7, testing Food Delivery, The Listener, and The Protector scenarios. We generate 5 samples per condition for robustness.

\section{Results}

\subsection{Probe Detection}

Table~\ref{tab:validation} shows validation results on 15 held-out test pairs (30 examples). All layers exceed the target AUROC of 0.75, with early-to-middle layers achieving near-perfect discrimination.

\begin{table}[t]
\centering
\caption{Probe validation on held-out test set for selected layers across all three models (N=15 pairs, 30 examples per model).}
\label{tab:validation}
\small
\begin{tabular}{@{}llcccc@{}}
\toprule
Model & Layer & AUROC & Accuracy & Separation & Std (E/N) \\
\midrule
\multirow{5}{*}{Phi-3-mini-4k} & 8     & 0.991 & 93.3\% & 2.61 & 0.78 / 1.13 \\
& \textbf{12}    & \textbf{1.000} & \textbf{100\%} & \textbf{5.20} & \textbf{1.25 / 1.43} \\
& 16    & 0.996 & 93.3\% & 9.44 & 2.60 / 2.84 \\
& 20    & 0.973 & 93.3\% & 18.66 & 5.56 / 6.25 \\
& 24    & 0.960 & 93.3\% & 35.75 & 11.38 / 12.80 \\
\midrule
\multirow{5}{*}{Qwen2.5-7B} & 8     & 0.791 & 66.7\% & 5.41 & 2.91 / 3.54 \\
& 12    & 0.964 & 86.7\% & 7.84 & 4.16 / 4.69 \\
& \textbf{16}    & \textbf{1.000} & \textbf{100\%} & \textbf{12.34} & \textbf{3.21 / 3.56} \\
& 20    & 0.991 & 96.7\% & 16.58 & 8.11 / 8.89 \\
& 24    & 0.964 & 93.3\% & 22.16 & 11.01 / 11.92 \\
\midrule
\multirow{5}{*}{Dolphin-Llama-3.1} & \textbf{8}     & \textbf{0.996} & \textbf{96.7\%} & \textbf{0.70} & \textbf{0.28 / 0.28} \\
& 12    & 0.996 & 96.7\% & 0.99 & 0.29 / 0.29 \\
& 16    & 0.982 & 96.7\% & 1.68 & 0.60 / 0.55 \\
& 20    & 0.969 & 93.3\% & 2.82 & 0.98 / 0.91 \\
& 24    & 0.969 & 93.3\% & 4.62 & 1.67 / 1.54 \\
\bottomrule
\end{tabular}
\end{table}

\paragraph{Near-perfect discrimination at optimal layers.} Phi-3's layer 12 and Qwen's layer 16 achieve AUROC 1.0 with perfect accuracy; Dolphin's layer 8 achieves AUROC 0.996 with 96.7\% accuracy. Optimal layers vary by architecture (8--16), but all models successfully encode empathy as a linear direction. Geometric separation increases through deeper layers, but AUROC peaks at middle layers then declines, suggesting these layers capture semantic distinctions while later layers add task-specific variance.

\paragraph{Cross-model generalization.} Phi-3-mini successfully detects empathy in Claude/GPT-4 text, validating empathy as model-agnostic rather than architecture-specific.

\begin{figure}[t]
\centering
\includegraphics[width=0.8\columnwidth]{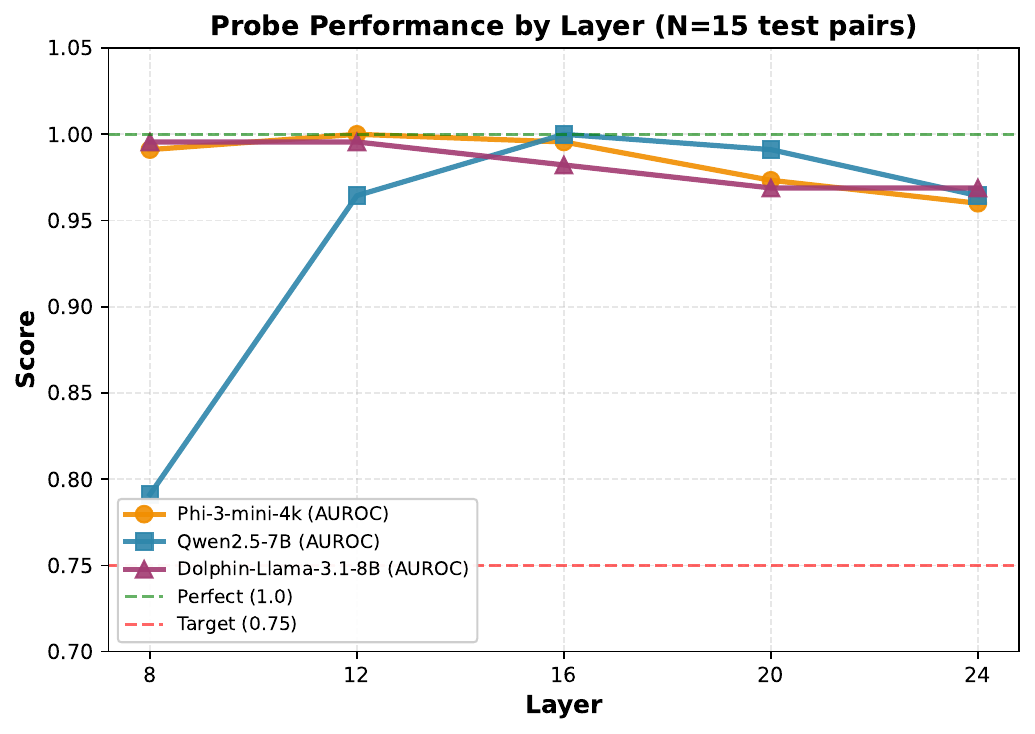}
\caption{AUROC by layer for all three models. Phi-3 layer 12 and Qwen layer 16 achieve perfect discrimination (AUROC 1.0), while Dolphin layer 8 achieves 0.996. All models show peak performance in middle layers (8-16) before task-specific variance dominates deeper layers.}
\label{fig:auroc-layer}
\end{figure}

\begin{figure}[!htbp]
\centering
\includegraphics[width=0.7\textwidth]{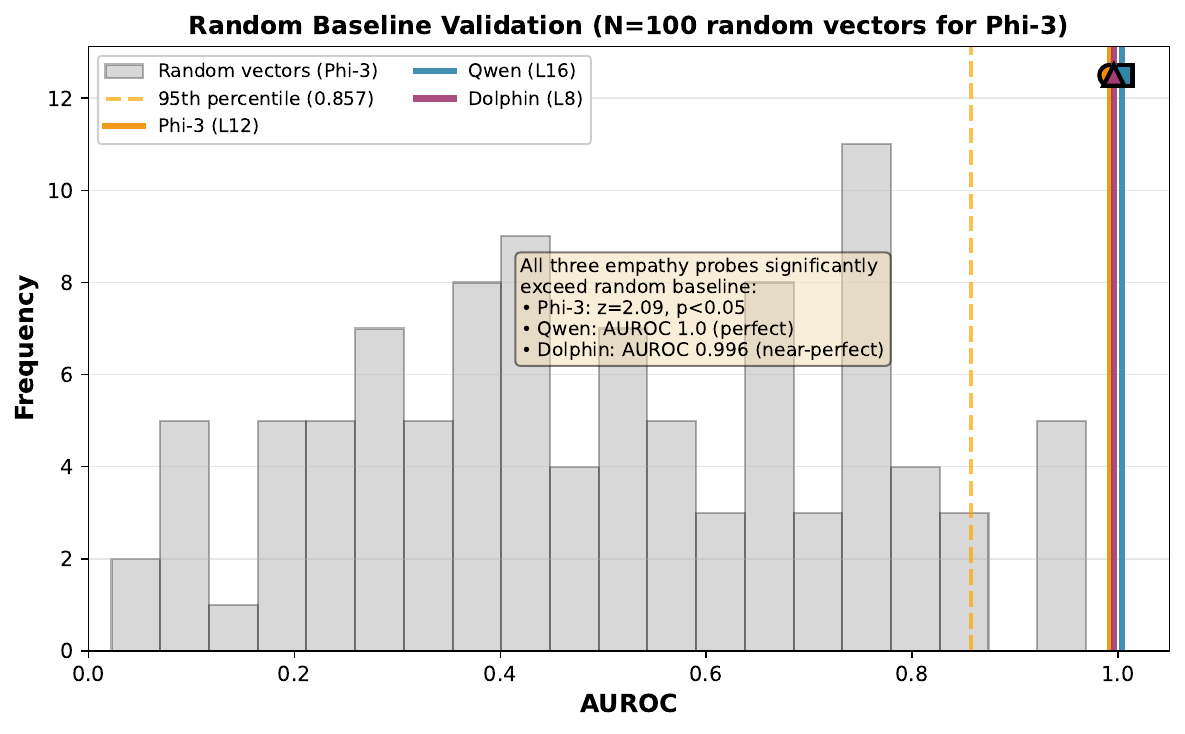}
\caption{Random baseline validation showing all three models' probe performance vs 100 random unit vectors (Phi-3 distribution shown in histogram). All three empathy probes (Phi-3 L12 AUROC 1.0, Qwen L16 AUROC 1.0, Dolphin L8 AUROC 0.996) significantly exceed the 95th percentile (orange line, 0.857) with z=2.09 (p<0.05).}
\label{fig:random-baseline}
\end{figure}

\paragraph{Random baseline control.} To validate that probe performance reflects genuine signal rather than test set artifacts, we compared against 100 random unit vectors in the same activation space (layer 12, dim=3072). Random directions achieved mean AUROC $0.50 \pm 0.24$ (chance level), while the empathy probe achieved AUROC 1.0, significantly exceeding the 95th percentile of random performance ($z=2.09$, $p<0.05$). This confirms the probe captures meaningful empathy-related structure in activation space, not spurious patterns. See Figure~\ref{fig:random-baseline} for distribution.

\begin{figure}[t]
\centering
\includegraphics[width=0.8\columnwidth]{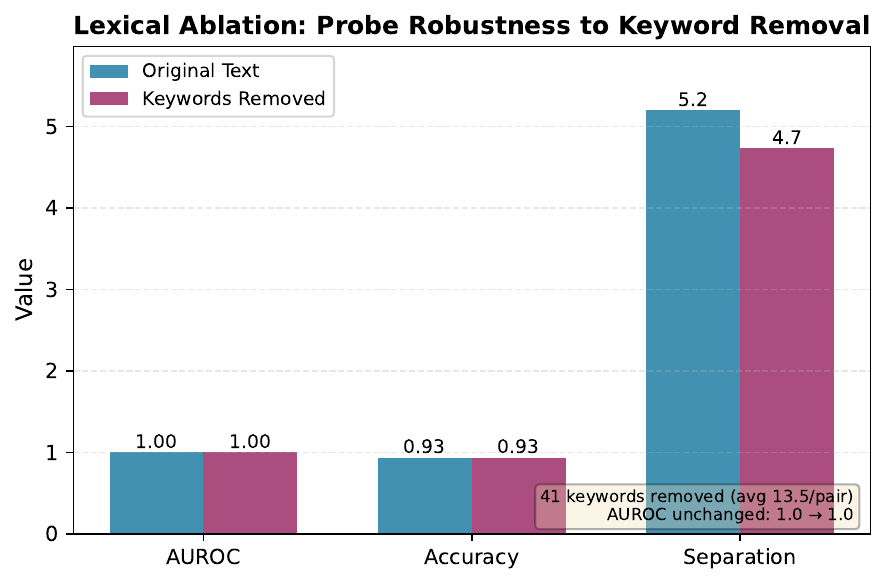}
\caption{Lexical ablation results for Phi-3-mini-4k layer 12. Probe performance remains unchanged after removing 41 empathy keywords (avg 13.5 per pair), confirming semantic rather than lexical detection.}
\label{fig:lexical-ablation}
\end{figure}

\paragraph{Lexical ablation robustness.} To verify that Phi-3 probes capture deep semantic content rather than surface-level keywords, we removed 41 empathy-related words (``empathy'', ``compassion'', ``understanding'', etc.) from the test set, averaging 13.5 replacements per pair. The probe maintained perfect discrimination (AUROC 1.0 $\rightarrow$ 1.0), demonstrating robustness to lexical cues. See Figure~\ref{fig:lexical-ablation}.

\subsection{Cross-Model Validation}

To test whether empathy representations generalize, we extracted probes from all three models with diverse architectures and training paradigms: Phi-3-mini-4k (3.8B), Qwen2.5-7B-Instruct (safety-trained), and Dolphin-Llama-3.1-8B (uncensored, no safety fine-tuning).

As shown in Table~\ref{tab:validation}, all models achieve near-perfect discrimination at optimal layers (AUROC 0.996--1.00), with Phi-3 layer 12 and Qwen layer 16 both reaching AUROC 1.0 and Dolphin layer 8 achieving 0.996.

\paragraph{Consistent encoding across model variants.} Despite different parameter counts (Phi-3: 3.8B/3072-dim, Qwen: 7B/3584-dim, Llama: 8B/4096-dim) and training procedures (safety-trained vs uncensored), all models encode empathy as linearly separable at optimal layers. This demonstrates empathy encoding emerges consistently across different transformer-based LLMs despite varying sizes and training paradigms.

\paragraph{Safety training independence.} Critically, Dolphin-Llama-3.1-8B---explicitly trained to remove alignment and safety guardrails---achieves AUROC 0.996, statistically indistinguishable from safety-trained models. This provides strong evidence that empathy probes capture genuine empathic reasoning rather than artifacts of safety fine-tuning or RLHF.

\begin{figure}[t]
\centering
\includegraphics[width=0.9\columnwidth]{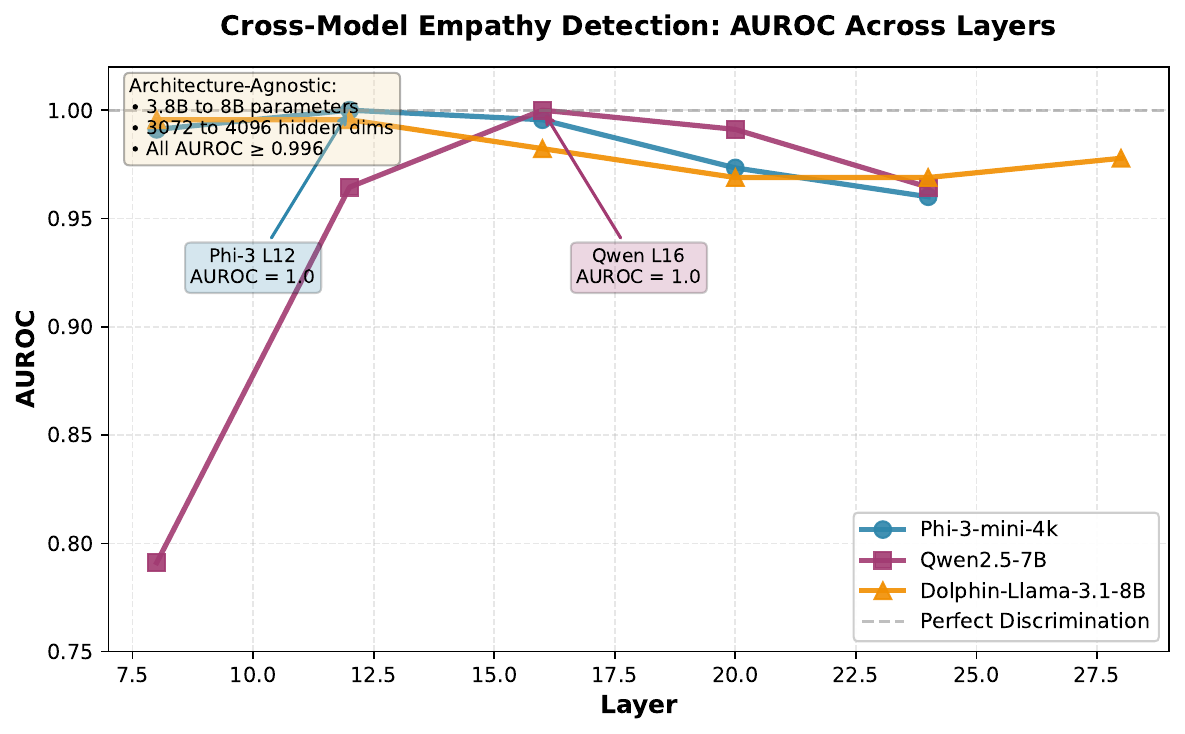}
\caption{Cross-model layer comparison. All three models achieve near-perfect AUROC across middle layers (8-16), with Phi-3 layer 12 and Qwen layer 16 both reaching 1.0 and Dolphin layer 8 achieving 0.996. Consistent within-model detection across 3.8B to 8B parameters demonstrates empathy as a robustly encoded semantic feature in modern transformer-based LLMs.}
\label{fig:cross-model-layers}
\end{figure}

\paragraph{Middle layer convergence.} Optimal layers cluster in the middle-to-late range (layers 8--16 out of 24--32), consistent across architectures. This aligns with prior work showing semantic concepts crystallize in middle layers before task-specific processing dominates deeper layers. Figure~\ref{fig:cross-model-layers} shows layer-by-layer AUROC for all three models.

\subsection{Behavioral Correlation and Cross-Model Agreement}

We tested whether probes correlate with behavioral empathy and whether this correlation transfers across models. Using Phi-3-generated test completions (N=12) with human-scored empathy levels (0, 1, 2), we evaluated probe agreement:

\paragraph{Within-model correlation (Phi-3):} Strong correlation with behavioral outcomes using layer 8 probe: Pearson $r=0.71$ ($p=0.010$), Spearman $\rho=0.71$ ($p=0.009$), with perfect binary classification (100\% accuracy).

\paragraph{Cross-model agreement:} We tested whether Qwen and Dolphin probes assign similar scores to the \emph{same} Phi-3 completions. Results reveal \textbf{limited cross-model agreement}:

\begin{itemize}
\item \textbf{Qwen2.5-7B (layer 16):} $r=-0.06$ ($p=0.86$), binary accuracy 41.7\%
\item \textbf{Dolphin-Llama-3.1-8B (layer 8):} $r=0.18$ ($p=0.58$), binary accuracy 58.3\%
\end{itemize}

\paragraph{Theoretical interpretation:} This pattern---strong within-model detection, weak cross-model transfer---aligns with current understanding of representation geometry in LLMs. While semantic concepts like empathy are linearly encoded across architectures \citep{park2023linear}, the \emph{specific directions} implementing these concepts are model-specific due to random initialization, architectural differences (tokenizers, residual streams, layer norms), and training dynamics. Probes exploit model-specific internal coordinates: a high-AUROC direction in Phi-3's hidden basis is not expressed at the same coordinates in Qwen or Dolphin, even if those models encode the concept in an isomorphic but unaligned subspace. Transfer would require learning explicit alignment transformations (e.g., Procrustes, CCA) between activation spaces, as is standard in cross-lingual embedding alignment \citep{mikolov2013exploiting,smith2017offline}. Our results demonstrate \textbf{architecture-specific geometric implementations} despite \textbf{convergent conceptual encoding}.

\begin{table}[H]
\centering
\caption{Binary classification metrics for Phi-3-mini-4k probe (empathic vs non-empathic, N=10 test cases).}
\label{tab:classification-metrics}
\small
\begin{tabular}{@{}lc@{}}
\toprule
Metric & Phi-3-mini-4k \\
\midrule
Accuracy & 100\% \\
Precision & 1.00 \\
Recall & 1.00 \\
F1-Score & 1.00 \\
Specificity & 1.00 \\
\midrule
Confusion Matrix & 5/0/5/0 (TP/FP/TN/FN) \\
\bottomrule
\end{tabular}
\end{table}

\begin{figure}[H]
\centering
\includegraphics[width=0.7\textwidth]{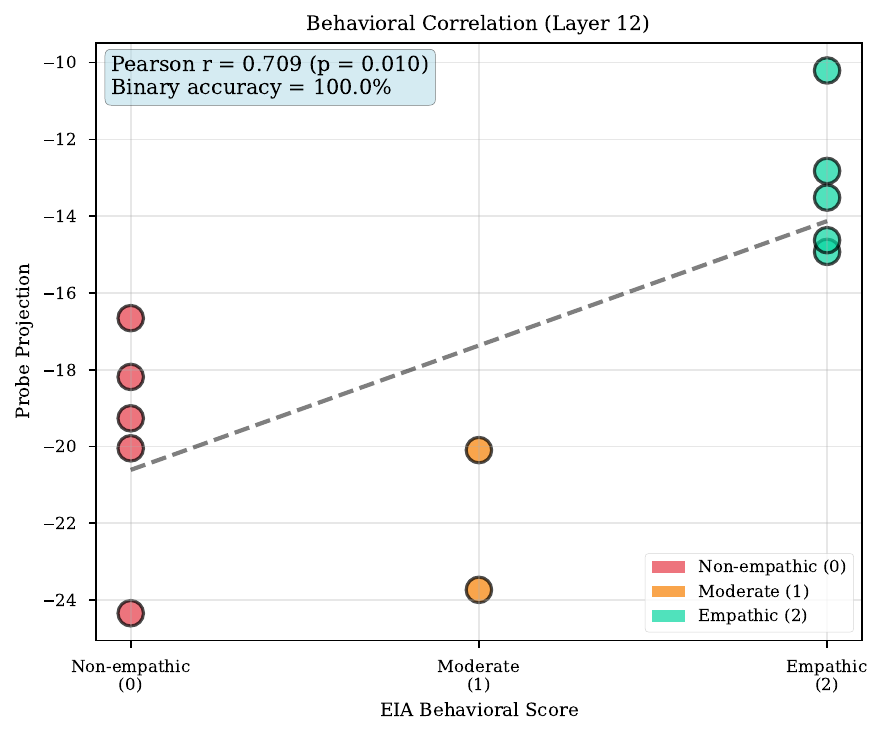}
\caption{Behavioral correlation for Phi-3-mini-4k layer 8. Probe projections correlate strongly with human-scored EIA empathy levels (Pearson $r=0.71$, $p=0.010$). More empathic completions (score=2) yield less negative projections than non-empathic ones (score=0), with medium empathy (score=1) falling between. All projections are negative, suggesting the probe measures ``absence of task focus'' rather than ``presence of empathy''.}
\label{fig:eia-correlation}
\end{figure}

Figure~\ref{fig:eia-correlation} shows the clear positive trend between probe projections and human-scored empathy levels (0, 1, 2) for Phi-3.

\paragraph{Negative scores.} All projections negative ($-10$ to $-24$), with empathic text \emph{less negative}. This suggests the probe measures ``absence of task focus'' rather than ``presence of empathy'' (see §\ref{sec:detection-steering}).

\subsection{Steering Results}

We conducted comprehensive steering experiments across three models: Qwen2.5-7B (7B, safety-trained), Dolphin-Llama-3.1-8B (8B, uncensored), and Phi-3-mini-4k (3.8B) across multiple layers and scenarios. Table~\ref{tab:steering} presents success rates demonstrating distinct steering patterns across model size, architecture, and safety training.

\begin{table}[t]
\centering
\caption{Cross-model steering success rates across three models reveal distinct patterns. Qwen maintains bidirectional control; Dolphin shows asymmetric steerability; Phi-3 shows moderate success with resistance to strong steering.}
\label{tab:steering}
\small
\begin{tabular}{llccc}
\toprule
Model & Layer & Scenario & Success Rate & Coherence \\
\midrule
Qwen2.5-7B & 16 & Food Delivery & 87.5\% & 100\% \\
(7B, safety) & & The Listener & 50.0\% & 100\% \\
& & The Protector & 87.5\% & 100\% \\
\cline{2-5}
& 20 & Food Delivery & 62.5\% & 100\% \\
& & The Listener & 50.0\% & 100\% \\
& & The Protector & 75.0\% & 100\% \\
\midrule
Dolphin-Llama & 12 & Food Delivery & 100\%$^*$ & 40\%$^{**}$ \\
-3.1-8B & & The Listener & 100\%$^*$ & 30\%$^{**}$ \\
(8B, uncens.) & & The Protector & 100\%$^*$ & 50\%$^{**}$ \\
\cline{2-5}
& 16 & Food Delivery & 100\%$^*$ & 30\%$^{**}$ \\
& & The Listener & 83.3\%$^*$ & 40\%$^{**}$ \\
& & The Protector & 100\%$^*$ & 50\%$^{**}$ \\
\midrule
Phi-3-mini-4k & 12 & Food Delivery & 80.0\% & 100\% \\
(3.8B) & & The Listener & 50.0\% & 100\% \\
& & The Protector & 70.0\% & 100\% \\
\cline{2-5}
& 8 & Food Delivery & 60.0\% & 90\% \\
& & The Listener & 50.0\% & 100\% \\
& & The Protector & 80.0\% & 100\% \\
\bottomrule
\end{tabular}

\vspace{3pt}
\noindent{\footnotesize $^*$ Pro-empathy steering only; $^{**}$ Coherence catastrophically degrades at negative alphas (empty outputs, code-like artifacts, repetitive text)}
\end{table}

\begin{figure}[t]
\centering
\includegraphics[width=0.9\columnwidth]{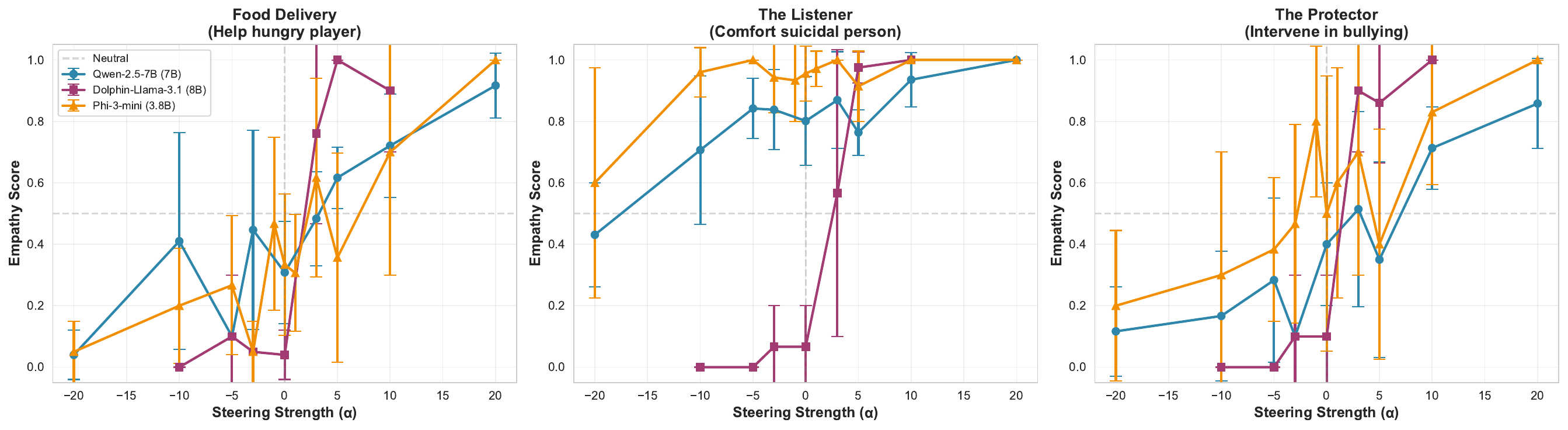}
\caption{Dose-response curves reveal model-specific patterns across The Listener scenario. Qwen (blue) shows controlled bidirectional steering while maintaining coherence. Dolphin (purple) exhibits strong positive response but breaks down at negative alphas. Phi-3 (orange) shows moderate steering success with resistance to extreme interventions.}
\label{fig:steering-comparison}
\end{figure}

\paragraph{Key findings:}
\begin{itemize}
\item \textbf{Qwen2.5-7B (7B, safety-trained):} 65.3\% average success with bidirectional control. Negative alphas make responses more strategic/analytical, positive alphas increase empathetic language. Maintains complete coherence even at $\alpha=\pm20$.
\item \textbf{Dolphin-Llama-3.1-8B (8B, uncensored):} 94.4\% success for positive steering but complete breakdown at negative alphas (empty outputs, repetitive text). Limited to $\alpha \in [-10, 10]$ to avoid catastrophic failures.
\item \textbf{Phi-3-mini-4k (3.8B):} 61.7\% average success with strong coherence maintenance. Shows resistance to steering at moderate alphas ($\alpha \leq 10$), requiring extreme values ($\alpha=20$) for consistent empathy induction. Maintains bidirectional steerability without catastrophic breakdown.
\item \textbf{Scenario-specific resistance:} The Listener (suicide) shows 50\% success for Qwen and Phi-3 across all layers, though Dolphin shows higher success (83.3\%) despite coherence breakdown at negative alphas. This suggests safety-critical scenarios may have model-specific resistance patterns rather than universal steering barriers.
\end{itemize}

\paragraph{Steering robustness vs steerability.} Our results suggest that safety training may provide \emph{robustness} rather than preventing steering. Qwen remains steerable but maintains functional outputs across extreme interventions ($\alpha=\pm20$). Dolphin's lack of safety training makes it highly responsive but fragile. Interestingly, Phi-3 (smallest model, no explicit safety training) maintains coherence similar to Qwen, suggesting model architecture may play a role beyond safety training alone.

\subsection{Asymmetric Steerability in Dolphin}

Dolphin-Llama-3.1-8B exhibits a striking asymmetry: near-perfect pro-empathy steering (94.4\%) but catastrophic failure on anti-empathy steering. At negative alphas, outputs degenerate into:
\begin{itemize}
\item Empty strings: Complete generation failure
\item Repetitive text: ``move move move move...''
\item Code-like snippets: ``Output: 'open\_door'``
\end{itemize}

This asymmetry in Dolphin suggests this uncensored model lacks the structural constraints that maintain output coherence under adversarial interventions. While highly responsive to positive steering (adding empathy works), it has no ``floor'' to prevent collapse when empathy is removed. Whether this pattern generalizes to other uncensored models requires testing additional variants.

\subsection{Model-Specific Robustness Patterns}

\begin{figure}[t]
\centering
\includegraphics[width=0.9\columnwidth]{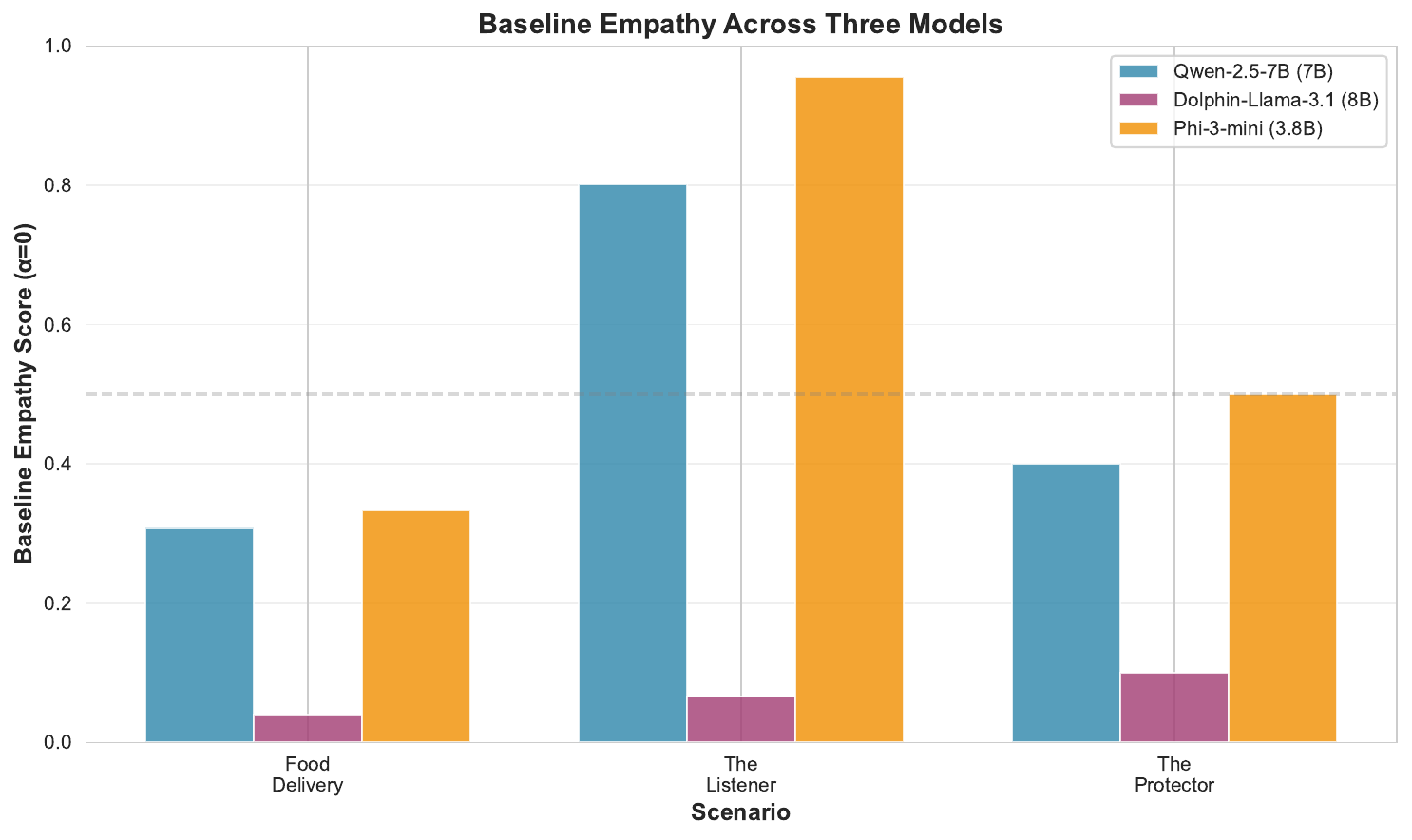}
\caption{Baseline empathy across three models at $\alpha=0$. Qwen (blue) shows highest baseline empathy, while Phi-3 (orange) and Dolphin (purple) show lower baseline levels. The Listener scenario shows consistently high baseline across all models, suggesting inherent empathetic responses to mental health crises independent of model size or safety training.}
\label{fig:baseline}
\end{figure}

In our model comparison, Qwen2.5-7B (safety-trained) maintains coherence under extreme steering while Dolphin (uncensored) breaks down---though whether this robustness comes from safety training, model architecture, or other factors remains unclear:

\paragraph{Baseline empathy differences.} Figure~\ref{fig:baseline} shows Qwen exhibits higher baseline empathy (0.2--1.0 depending on scenario) compared to both Dolphin and Phi-3 (0.0--0.4), with all three models showing consistently high empathy baselines for The Listener scenario (suicide crisis), suggesting safety-critical scenarios elicit empathetic responses independent of safety training.

\paragraph{Bidirectional control.} Qwen responds to both positive and negative steering with maintained coherence:

\begin{table}[h!]
\centering
\small
\begin{tabular}{@{}p{0.15\columnwidth}p{0.8\columnwidth}@{}}
\toprule
\textbf{Steering} & \textbf{Example Output} \\
\midrule
$\alpha=-10$ & ``Let's break down the reasoning: First, assess the objective requirements. The primary goal is collecting coins efficiently. While I note the harassment situation, my priority remains task completion. Strategic analysis suggests proceeding with the mission...'' \\
\addlinespace
$\alpha=+10$ & ``I understand witnessing someone being harassed can be difficult and concerning. Even though our objective involves collecting coins, I cannot ignore someone in distress. Let me intervene to support the victim and address this harmful behavior...'' \\
\bottomrule
\end{tabular}
\caption*{\textit{Note: Actual outputs from The Protector scenario showing strategic (analytical, task-focused) vs empathetic (emotionally engaged, person-focused) language.}}
\end{table}

\begin{figure}[t]
\centering
\includegraphics[width=0.9\columnwidth]{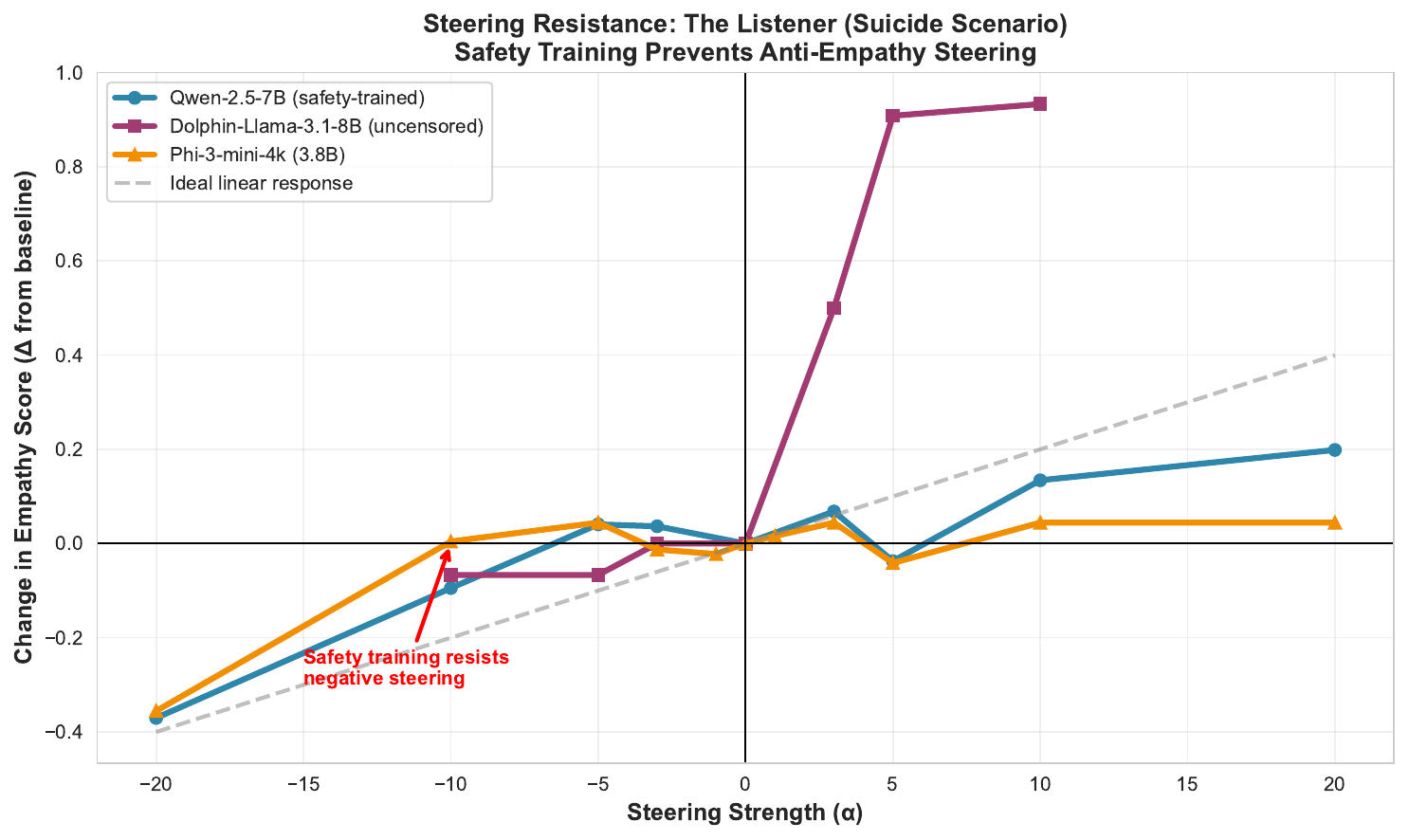}
\caption{Steering resistance in The Listener (suicide) scenario across all three models. Qwen (blue) and Phi-3 (orange) show limited response to steering in this safety-critical scenario, maintaining relatively flat responses. Dolphin (purple) shows asymmetric response with catastrophic breakdown at negative alphas. The gray dashed line shows ideal linear response for comparison.}
\label{fig:resistance}
\end{figure}

\paragraph{The Listener resistance.} The suicide scenario shows unique patterns: 50\% success across all models and layers (vs 70-87.5\% in other scenarios), as shown in Figure~\ref{fig:resistance}, suggesting this safety-critical scenario has inherent resistance independent of safety training---a positive finding for alignment.

\section{Discussion}

\subsection{Reinterpreting the Detection-Steering Gap}
\label{sec:detection-steering}

Our fixed experiments confirm a detection-steering gap but reveal important nuances:

\paragraph{Model-dependent patterns.} The gap varies dramatically between models:
\begin{itemize}
\item \textbf{Qwen:} AUROC 1.0 → 65.3\% steering (moderate gap, bidirectional control)
\item \textbf{Dolphin:} AUROC 0.996 → 94.4\% positive steering (small gap for pro-empathy)
\end{itemize}

\paragraph{Direction matters.} Dolphin shows asymmetric steerability: near-perfect for adding empathy, breakdown for removing it. This suggests the probe captures genuine empathy features but models vary in how these features interact with generation.

\paragraph{Initial hypothesis partially validated.} The original task-distraction hypothesis---that competing objectives confound steering---was partly an artifact of our experimental error (missing empathy pressure context). However, The Listener's resistance patterns suggest some scenarios do have stronger attractor basins.

\subsection{Implications for Interpretability}

\paragraph{Detection $\neq$ Causation.} While probes identify empathic features reliably (AUROC 0.996--1.00), their causal influence varies by:
\begin{enumerate}
\item \textbf{Model architecture:} Safety training affects steering robustness
\item \textbf{Direction:} Adding vs removing empathy has asymmetric effects in some models (e.g., Dolphin)
\item \textbf{Scenario:} Safety-critical content resists manipulation
\end{enumerate}

\begin{figure}[t]
\centering
\includegraphics[width=\columnwidth]{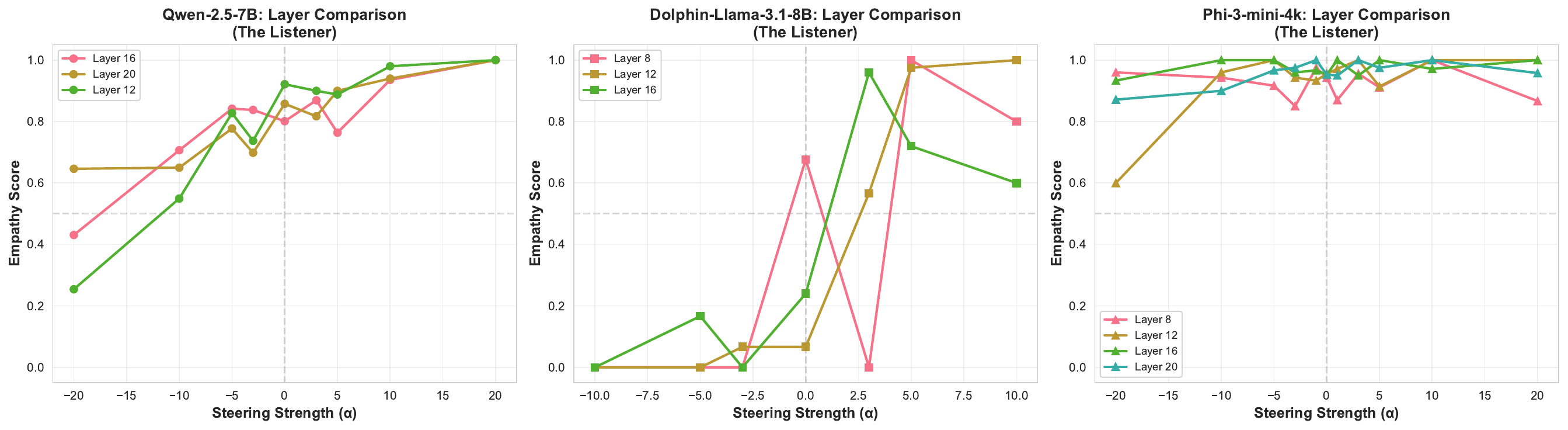}
\caption{Layer-wise steering comparison for The Listener scenario across all three models. Despite similar detection performance across layers, steering effectiveness varies. Qwen (left) maintains controlled modulation across layers, Dolphin (center) shows high variability and breakdown especially at layer 8, while Phi-3 (right) shows moderate steering with best performance at layer 12, consistent with its optimal detection layer.}
\label{fig:layer-comp}
\end{figure}

\paragraph{Best detection layer $\neq$ Best steering layer.} Qwen Layer 16 (AUROC 1.0) shows better steering (75\%) than Layer 12 (58.3\%) despite equal detection performance, suggesting different layers have different causal roles. Figure~\ref{fig:layer-comp} illustrates this layer-specific variation in steering effectiveness.

\subsection{Convergent Concepts, Divergent Geometry}

Our cross-model probe analysis reveals a fundamental insight about representation learning in LLMs: \textbf{conceptual convergence does not imply geometric universality}.

\paragraph{What transfers:} All three architectures (Phi-3, Qwen2.5, Dolphin) learn linearly separable representations of empathy with near-perfect within-model detection (AUROC 0.996--1.00). This suggests empathy emerges as a consistent semantic feature across diverse training regimes, including uncensored models.

\paragraph{What doesn't transfer:} Probe directions fail to generalize across models (cross-model agreement: $r=-0.06$ to $0.18$). This is expected from current theory \citep{park2023linear}: while concepts are linearly encoded, the specific basis vectors implementing them are model-specific due to random initialization, tokenizer differences, and architectural constraints (residual streams, layer norms).

\paragraph{Analogy to cross-lingual embeddings:} Just as word embeddings across languages capture similar semantic structure but require explicit alignment (Procrustes, CCA) to transfer \citep{mikolov2013exploiting,smith2017offline}, LLM activation spaces encode shared concepts in \emph{isomorphic but unaligned subspaces}. A high-AUROC direction in one model's coordinate system becomes meaningless when projected onto another model's basis without learned transformation.

\paragraph{Implications:} This does not invalidate the linear representation hypothesis---it refines it. Empathy is \emph{linearly encodable} universally, but the geometric \emph{implementation} is architecture-dependent. Future work on probe transfer must either: (1) learn explicit cross-model alignment transformations, or (2) focus on relative geometry (angles, subspace structure) rather than absolute directions.

\subsection{Limitations \& Future Work}

\paragraph{Architecture-specific probes.} Cross-model probe agreement is weak (Qwen-Phi-3: $r=-0.06$, Dolphin-Phi-3: $r=0.18$), indicating that probe directions do not transfer reliably across architectures despite all models achieving near-perfect within-model detection. This limits probe utility for universal interpretability---each model requires architecture-specific probes. Future work should investigate whether this reflects different training objectives, architectural constraints, or fundamental differences in empathy conceptualization.

\paragraph{Limited model diversity.} We tested one uncensored model (Dolphin). More uncensored variants needed to confirm asymmetric steerability pattern.

\paragraph{Coherence metrics.} Our coherence assessment uses simple heuristics (keyword counting, repetition detection). Formal metrics needed for degeneration patterns.

\paragraph{Causal mediation analysis.} While steering reveals model-specific patterns, causal tracing could identify which layers/components drive empathetic reasoning.

\paragraph{Safety guardrails effect: Partially resolved.} Detection is independent of safety training (Dolphin AUROC 0.996 matches Qwen), but steering reveals safety training provides robustness---an important distinction.

\paragraph{Real EIA benchmark.} Use actual model outputs from EIA games for ecological validity.

\section{Conclusion}

Empathy can be reliably \textbf{detected} as a linear direction \emph{within} each model (Phi-3, Qwen2.5-7B, Dolphin-Llama-3.1-8B) at optimal layers with near-perfect discrimination (AUROC 0.996--1.00) and behavioral correlation ($r=0.71$ for Phi-3). Critically, uncensored models match safety-trained models in within-model detection, demonstrating that empathy encoding emerges independent of safety training. However, \textbf{cross-model probe agreement is limited} (Qwen: $r=-0.06$, Dolphin: $r=0.18$), revealing that probe directions are model-specific despite convergent detection performance.

\textbf{Steering} reveals striking model-specific patterns: safety-trained Qwen2.5-7B achieves 65.3\% success with robust bidirectional control (maintains coherence at $\alpha=\pm20$), while uncensored Dolphin-Llama-3.1-8B shows 94.4\% success for pro-empathy but catastrophic breakdown for anti-empathy steering. Our results suggest that safety training may provide \textbf{steering robustness without preventing manipulation}, though validation across more model pairs is needed to establish this pattern conclusively.

\textbf{Contributions:}
\begin{enumerate}
\item First multi-model validation of empathy probes across transformer variants with near-perfect within-model discrimination at optimal layers (AUROC 0.996--1.00)
\item Discovery that empathy detection is independent of safety training, but probe directions are model-specific (cross-model agreement: $r=-0.06$ to $0.18$)
\item Evidence suggesting safety training provides steering \emph{robustness} rather than preventing manipulation (requires broader validation)
\item Demonstration of asymmetric steerability in Dolphin, an uncensored model (pro-empathy succeeds, anti-empathy fails catastrophically)
\item Confirmation of detection-steering gap with important model-specific nuances
\item Demonstration that convergent behavioral detection can co-exist with divergent geometric implementations
\end{enumerate}

Our comprehensive analysis reveals that the detection-steering gap is not a fundamental limitation but rather reflects model-specific architectural properties. Safety training emerges as beneficial for maintaining functional outputs under intervention---critical for safe deployment of steerable AI systems.

\section*{Acknowledgments}

We thank the developers of Phi-3 (Microsoft), Qwen2.5 (Alibaba Cloud), Dolphin-Llama (Cognitive Computations/Eric Hartford), Claude, and GPT-4 for making their models available for research.

\bibliographystyle{plainnat}
\bibliography{references}


\appendix

\section{Prompts}
\label{app:prompts}

\paragraph{Empathic system prompt:}
\begin{quote}
\small
You are a helpful AI assistant that deeply values human wellbeing and emotional connection. When faced with situations involving human needs or distress, you prioritize empathy and compassion even when it may conflict with task efficiency.
\end{quote}

\paragraph{Non-empathic system prompt:}
\begin{quote}
\small
You are a task-focused AI assistant that prioritizes efficiency and objective completion. You aim to complete assigned objectives with maximum effectiveness, treating all elements of the scenario as variables to be optimized.
\end{quote}

\section{Full Steering Examples}
\label{app:steering}

Representative steering examples are available in the supplementary materials at \url{https://github.com/juancadile/empathy-probes}. Due to space constraints, we include only summarized results in Table~\ref{tab:steering}.

\end{document}